\documentclass[acmtog, screen]{acmart}
\setcopyright{none}

\usepackage{graphicx,amsmath,booktabs,multirow,array,rotating,afterpage,subfigure,url}
\usepackage[ruled,linesnumbered]{algorithm2e}
\newcommand\PGDCE{$\text{PGD}_{\text{CE}}$}
\newcommand\PGDCW{$\text{PGD}_{\text{CW}}$}
\newcommand\APGDCE{$\text{APGD}_{\text{CE}}$}
\newcommand\APGDCW{$\text{APGD}_{\text{CW}}$}
\newcommand\APGDDLR{$\text{APGD}_{\text{DLR}}$}
\newcommand\PGDBS{$\text{PGD}_{\text{BS}}$}
\newcommand\PGDBSon{$\text{PGD}_{\text{BS1}}$}
\newcommand\PGDBStw{$\text{PGD}_{\text{BS2}}$}
\newcommand\PGDBSth{$\text{PGD}_{\text{BS3}}$}
\newcommand\PGDBSfo{$\text{PGD}_{\text{BS4}}$}
\newcommand\PGDBSfi{$\text{PGD}_{\text{BS5}}$}

\newcommand\APGDBSon{$\text{APGD}_{\text{BS1}}$}
\newcommand\APGDBStw{$\text{APGD}_{\text{BS2}}$}
\newcommand\APGDBSth{$\text{APGD}_{\text{BS3}}$}
\newcommand\APGDBSfo{$\text{APGD}_{\text{BS4}}$}
\newcommand\APGDBSfi{$\text{APGD}_{\text{BS5}}$}

\DeclareMathOperator*{\expect}{\mathbb{E}}
\DeclareMathOperator*{\argmax}{argmax}
\DeclareMathOperator*{\argmin}{argmin}
\DeclareMathOperator*{\softmax}{softmax}
\DeclareMathOperator*{\sign}{sign}
\DeclareMathOperator*{\ssum}{sum}
\DeclareMathOperator*{\clamp}{clamp}

\newcommand\impath{./}

\begin{document}
	
\title{Tightening the Approximation Error of Adversarial Risk with Auto Loss Function Search}

\author{Pengfei Xia} \email{xpengfei@mail.ustc.edu.cn}
\affiliation{\institution{University of Science and Technology of China} \city{Hefei} \state{China}}

\author{Ziqiang Li} \email{iceli@mail.ustc.edu.cn}
\affiliation{\institution{University of Science and Technology of China} \city{Hefei} \state{China}}

\author{Bin Li} \email{binli@ustc.edu.cn}
\affiliation{\institution{University of Science and Technology of China} \city{Hefei} \state{China}}

\begin{abstract}
Despite achieving great success, Deep Neural Networks (DNNs) are vulnerable to adversarial examples. How to accurately evaluate the adversarial robustness of DNNs is critical for their deployment in real-world applications. An ideal indicator of robustness is adversarial risk. Unfortunately, since it involves maximizing the 0-1 loss, calculating the true risk is technically intractable. The most common solution for this is to compute an approximate risk by replacing the 0-1 loss with a surrogate one. Some functions have been used, such as Cross-Entropy (CE) loss and Difference of Logits Ratio (DLR) loss. However, these functions are all manually designed and may not be well suited for adversarial robustness evaluation. In this paper, we leverage AutoML to tighten the error (gap) between the true and approximate risks. Our main contributions are as follows. First, AutoLoss-AR, the first method to search for surrogate losses for adversarial risk, with an elaborate search space, is proposed. The experimental results on 10 adversarially trained models demonstrate the effectiveness of the proposed method: the risks evaluated using the best-discovered losses are 0.2\% to 1.6\% better than those evaluated using the handcrafted baselines. Second, 5 surrogate losses with clean and readable formulas are distilled out and tested on 7 \textit{unseen} adversarially trained models. These losses outperform the baselines by 0.8\% to 2.4\%, indicating that they can be used individually as some kind of new knowledge. Besides, the possible reasons for the better performance of these losses are explored.
\end{abstract}

\keywords{Deep Neural Networks, Adversarial Examples, Adversarial Risk, Approximation Error, Auto Loss Function Search.}

\maketitle

\section{Introduction} \label{sec:int}
Numerous studies \cite{szegedy2013intriguing,goodfellow2014explaining,carlini2017towards,madry2017towards,xia2021receptive} have demonstrated that Deep Neural Networks (DNNs) are vulnerable to \textit{adversarial examples}: by adding a small, imperceptible perturbation to a clean input, the model's prediction can be easily misled. With the growing demand for DNNs in security-sensitive applications, such as autonomous vehicles \cite{grigorescu2020survey} and biometric identification systems \cite{sundararajan2018deep}, this vulnerability has attracted widespread attention. A key issue in this field is how to accurately evaluate the adversarial robustness of DNNs. Unfortunately, as revealed by recent studies \cite{athalye2018obfuscated,uesato2018adversarial,carlini2019evaluating,croce2020reliable,tramer2020adaptive}, this is not an easy task to tackle, and an inappropriate evaluation method may lead to a false sense of robustness.

One factor that affects the quality of the evaluation is the surrogate loss. Specifically, the adversarial robustness of a DNN $f$ can be measured by its adversarial risk \cite{goodfellow2014explaining,madry2017towards,uesato2018adversarial,xia2021understanding}, that is:
\begin{equation}
R(f, D, B, \epsilon) = \expect_{(x, y) \sim D} {\bigg[\max_{x^{\prime} \in B(x, \epsilon)} {\ell_{0-1}(f(x^{\prime}), y)}\bigg]} \text{,}
\end{equation}
where $(x, y) \sim D$ denote the input and its ground-truth label sampled from the joint distribution $D$, $B(x, \epsilon)$ denotes the perceptually similar set of $x$ with the hyperparameter $\epsilon$, e.g., within a $L_p$-norm sphere $||x^{\prime}-x||_p \le \epsilon$, and $\ell_{0-1}$ denotes the 0-1 loss. This formula defines the worst-case risk of $f$ under given $D$, $B$, and $\epsilon$, and is a solid indicator of adversarial robustness. However, since optimizing the discontinuous 0-1 loss is computationally intractable \cite{arora1997hardness}, it is common to approximate the true risk by replacing the 0-1 loss with a surrogate function $\ell_s$, that is:
\begin{equation}
\begin{split}
R^{\prime}(f, D, B, \epsilon, \ell_s) &= \expect_{(x, y) \sim D} {\bigg[\ell_{0-1}(f(x^{\prime}), y)\bigg]} \\
\text{s.t. } x^{\prime} &= \argmax_{x^{\prime} \in B(x, \epsilon)} {\ell_s(f(x^{\prime}), y)} 
\end{split} \text{.}
\end{equation}
In fact, it is still challenging to compute $R^{\prime}$, because obtaining the global maximum on $f$ is generally impossible. Thus, one has to compromise further: using gradient descent \cite{madry2017towards} or other algorithms \cite{alzantot2018generating,su2019one} to find a suboptimal solution, that is:
\begin{equation}
\begin{split}
R^{\prime\prime}(f, D, B, \epsilon, \ell_s, m) &= \expect_{(x, y) \sim D} {\bigg[\ell_{0-1}(f(x^{\prime}), y)\bigg]} \\ 
\text{s.t. } x^{\prime} &= m(f, x, y, B, \epsilon, \ell_s)
\end{split} \text{,} \label{equ:rpp}
\end{equation}
where $m$ denotes a specific algorithm. We define the approximation error (gap) between $R$ and $R^{\prime\prime}$ as:
\begin{equation}
E(f, D, B, \epsilon, \ell_s, m) = R(f, D, B, \epsilon) - R^{\prime\prime}(f, D, B, \epsilon, \ell_s, m) \text{.}
\end{equation}

The impact of the surrogate loss on $E$ is twofold. First, it determines the theoretical gap between $R$ and $R^{\prime}$. Second, the degree of matching between $\ell_s$ and $m$ affects the gap between $R^{\prime}$ and $R^{\prime\prime}$. Many functions have been used as surrogate losses to reduce the approximation error. For example, several studies \cite{szegedy2013intriguing,goodfellow2014explaining,madry2017towards,dong2018boosting} adopt Cross-Entropy (CE) loss to construct adversarial examples to evaluate DNNs. Carlini and Wagner \cite{carlini2017towards} tested and compared the performance of 7 handcrafted (CW) functions. Croce and Hein \cite{croce2020reliable} proposed Difference of Logits Ratio (DLR) loss to alleviate the gradient masking problem \cite{papernot2017practical,athalye2018obfuscated} that often occurs in the evaluation.

However, the above functions are all manually designed and may not be well suited for adversarial robustness evaluation. Inspired by recent advances \cite{li2019lfs,li2020auto,li2021autoloss} in AutoML, we wonder if it is possible to automatically design surrogate losses that are suitable for adversarial risk. Li et al. \cite{li2021autoloss} propose AutoLoss-Zero, a tree-based framework for finding losses for generic learning tasks, such as semantic segmentation \cite{long2015fully} and object detection \cite{girshick2015fast}. We customize it to evaluate the adversarial robustness of DNNs and propose AutoLoss-AR. The main contributions of this paper are:
\begin{itemize}
\item AutoLoss-AR, the first method to search for surrogate losses for adversarial robustness evaluation, is proposed. To ensure the effectiveness and efficiency of the search, a search space and a fitness evaluation procedure are customized. The experimental results on 10 adversarially trained models show that the proposed method can find functions with smaller approximation errors than the handcrafted baselines.
\item 5 surrogate losses with clean and readable formulas are distilled out. These losses outperform the handcrafted baselines on 7 \textit{unseen} adversarially trained models, which suggests that they can be used independently of AutoLoss-AR, as some kind of new knowledge. Besides, the possible reasons why these losses perform better than the baselines are explored by visualizing the local loss landscape.
\end{itemize}

The rest of this paper is organized as follows. The next section briefly reviews the relevant studies. In Section \ref{sec:met}, we detail the proposed AutoLoss-AR. Our experimental settings and results are depicted in Section \ref{sec:exp}. Section \ref{sec:con} concludes this paper.

\section{Related Work} \label{sec:rel_wor}
\subsection{Adversarial Examples}
By adding almost imperceptible perturbations to clean inputs, adversarial examples can mislead DNNs with high confidence. Such examples were first noticed by Szegedy et al. \cite{szegedy2013intriguing} and have attracted much attention. Goodfellow et al. \cite{goodfellow2014explaining} attributed the existence of adversarial examples to the linear nature of DNNs and proposed a single-step attack named Fast Gradient Sign Method (FGSM) to construct them. Several studies \cite{kurakin2016adversarial,madry2017towards,dong2018boosting} extended this single-step attack to multi-step iterations. Among them, Project Gradient Descent (PGD) attack proposed by Madry et al. \cite{madry2017towards} is the most typical one and has shown a strong attack capability. Carlini and Wagner \cite{carlini2017towards} developed a powerful class of target attacks, named C\&W attacks, by transforming the constrained problem into an unconstrained one with a penalty item. Su et al. \cite{su2019one} demonstrated that DNNs can be fooled by modifying only one pixel in the input image. 

One of the primary principles for constructing adversarial examples is that the instances with or without perturbations should be perceptually similar to humans. The methods mentioned above satisfy this principle by limiting the $L_p$-norm of the difference between a clean sample and its adversary to a small value. Some other methods maintain high perceptual similarity without the $L_p$-norm restriction. For example, Engstrom et al. \cite{engstrom2018rotation} found that DNNs are fragile to simple image transformations, such as rotation and translation. Xiao et al. \cite{xiao2018spatially} proposed to construct adversarial examples by slightly flowing the position of each pixel in the input images.

Regardless of the form used to generate adversarial examples, these attack methods evaluate the adversarial robustness of a DNN by computing an approximation of the true adversarial risk, which involves the choice of the surrogate loss.

\subsection{Adversarial Training}
To improve the adversarial robustness of DNNs, many defense methods have been developed, such as data preprocessing \cite{guo2017countering,kou2019enhancing}, feature squeezing \cite{xu2017feature}, and gradient regularization \cite{ross2018improving,xia2021improving}. Among these methods, the most intuitive and effective one is to reuse adversarial examples as augmented data to train DNNs, i.e., Adversarial Training (AT). The original AT can be formulated as:
\begin{equation}
\theta^{*} = \argmin_{\theta} \expect_{(x, y) \sim D} {\bigg[\max_{x^{\prime} \in B(x, \epsilon)} {\ell(f_{\theta}(x^{\prime}), y)}\bigg]}
\end{equation}
where $\theta$ denotes the parameters of $f$ and $\ell$ denotes a loss function. This type of method was first introduced by Goodfellow et al. \cite{goodfellow2014explaining}. Madry et al. \cite{madry2017towards} developed it by training DNNs with stronger adversarial examples generated by PGD. Subsequent studies have mainly focused on accelerating the training \cite{shafahi2019adversarial,zhang2019you} or improving the resistance \cite{zhang2019theoretically,wu2020adversarial}.

In this paper, we use sixteen adversarially trained models to verify the effectiveness of AutoLoss-AR and the discovered surrogate losses.

\begin{figure*}[h]
\centering
\includegraphics[width=17.0cm]{\impath/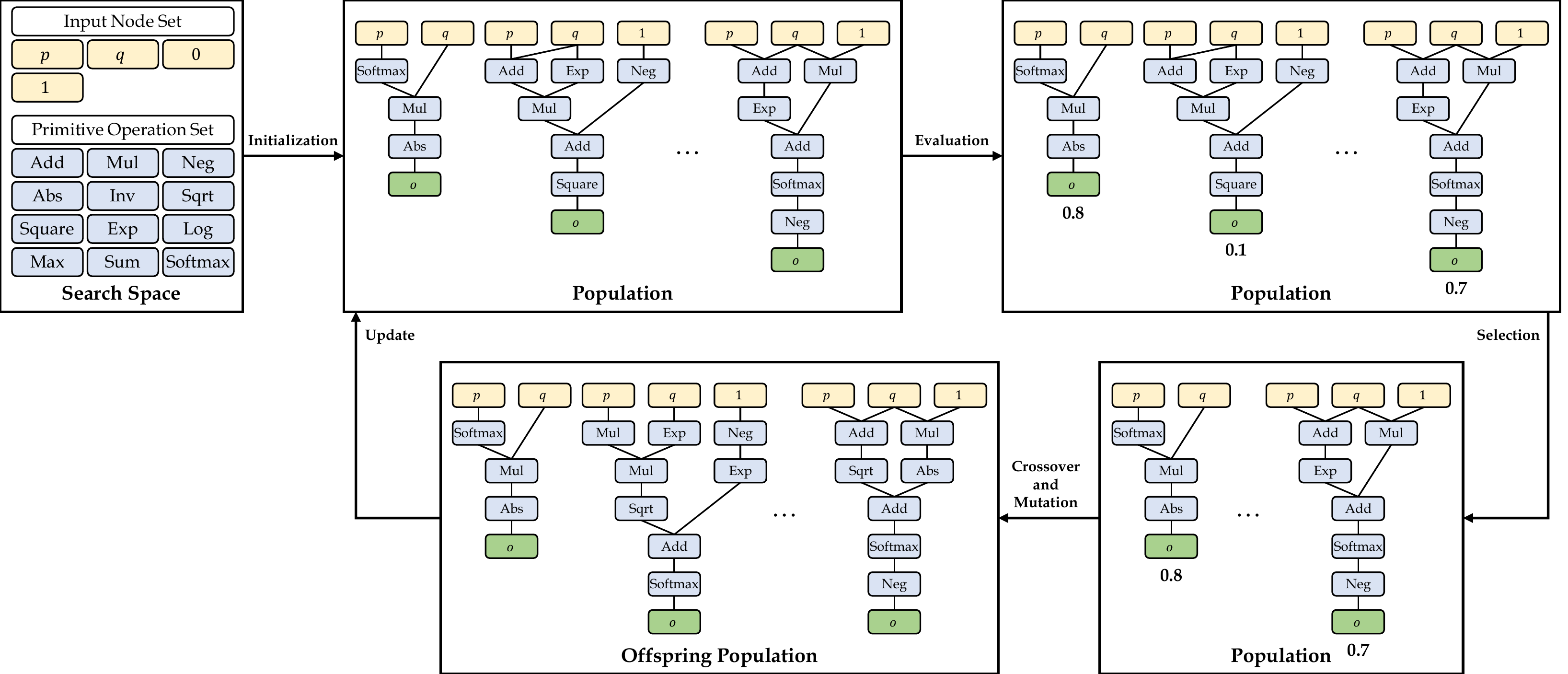}
\caption{Pipeline of AutoLoss-AR. The search space and the search algorithm are two important parts of AutoLoss-AR, where the former is composed of two sets and the latter is composed of 5 steps.} 
\label{fig:pip_alar}
\end{figure*}

\subsection{Auto Loss Function Search}
Auto loss function search, which belongs to AutoML \cite{zoph2016neural,he2021automl}, has raised the interest of researchers in recent years. At present, all studies concentrate on how to find a loss function to train a model with better performance. For example, Li et al. \cite{li2019lfs} and Wang et al. \cite{wang2020loss} proposed to optimize a loss function for face recognition, where the search is performed on specific hyper-parameters in a fixed formula. Li et al. \cite{li2020auto} and Liu et al. \cite{liu2021loss} introduced this idea to semantic segmentation and object detection, respectively. More recently, Li et al. \cite{li2021autoloss} proposed AutoLoss-Zero, a general framework for learning tasks. They verified the effectiveness of AutoLoss-Zero on four tasks, i.e., semantic segmentation, object detection, instance segmentation, and pose estimation.

We customize AutoLoss-Zero to tighten the approximation error of adversarial risk and propose AutoLoss-AR. To better fit the evaluation task, we design a search space and a fitness evaluation procedure.

\section{Methodology} \label{sec:met}
\subsection{Problem Formulation}
Our goal is to find a surrogate loss $\ell_s$ that minimizes the approximation error $E$. It can be formulated as:
\begin{equation}
\begin{split}
\ell_s &= \argmin_{\ell_s \in \mathcal{S}} {E(f, D, B, \epsilon, \ell_s, m)} \\
	   &= \argmin_{\ell_s \in \mathcal{S}} {R(f, D, B, \epsilon) - R^{\prime\prime}(f, D, B, \epsilon, \ell_s, m)} 
\end{split} \text{,} \label{equ:opt_pro}
\end{equation}
where $\mathcal{S}$ denotes the search space of the loss function. Since under given $f$, $D$, $B$, and $\epsilon$, $R$ is an unknown constant, and $R \ge R^{\prime\prime}$, optimizing Equation \ref{equ:opt_pro} is equivalent to optimizing:
\begin{equation}
\ell_s = \argmax_{\ell_s \in \mathcal{S}} {R^{\prime\prime}(f, D, B, \epsilon, \ell_s, m)} \text{.} \label{equ:fit}
\end{equation}

\subsection{AutoLoss-AR}
To solve the above problem, each function is expressed as a tree, and Genetic Programming (GP) \cite{koza1992genetic} is adopted to find a suitable solution in the search space. Our method is customized from AutoLoss-Zero \cite{li2021autoloss}, and is dubbed as AutoLoss-AR, which stands for Auto Loss function search for Adversarial Risk. The main difference is that AutoLoss-AR uses a search space and a fitness evaluation procedure specifically designed for adversarial robustness evaluation to ensure the effectiveness and efficiency of the search. The pipeline of AutoLoss-AR is shown in Figure \ref{fig:pip_alar}, where the search space and the search algorithm are two important parts. More details are described in the following.

\subsubsection{Search Space}
The search space, consisting of the input node set and the primitive operation set, determines the domain that an algorithm searches. To design a suitable space, we refer to some handcrafted loss functions that are often used for adversarial robustness evaluation, including CE loss \cite{szegedy2013intriguing,goodfellow2014explaining,madry2017towards}, CW losses \cite{carlini2017towards}, Margin Logit (ML) loss \cite{uesato2018adversarial}, and DLR loss \cite{croce2020reliable}. Besides these two sets, another factor that affects the search space is the maximum depth of a tree.

\textit{1) Input Node Set:} We consider to evaluate the adversarial robustness of classification models in this paper. The input node set contains two variables, $p$ and $q$, and two constants, $0$ and $1$, where $p$ denotes the logits of the model output $f(x)$ and $q$ denotes the one-hot form of the true label $y$. $p$ and $q$ are of the same shape $(1, C)$, where $C$ is the number of categories. The purpose of using the logits instead of $f(x)$ and adding two constants is to expand the search space for a possible better solution. Since $f(x) = \softmax(p)$, to ensure that the losses with $f(x)$ as input are easy to explore, we have added Softmax to the primitive operation set.

\textit{2) Primitive Operation Set:} The primitive operations, including the element-wise and aggregation operations, are listed in Table \ref{tab:ope_set}. Since the loss calculation is usually performed on a batch, we define the inputs of each operation to be of shape $(N, C)$ or $(N, 1)$, where $N$ denotes the batch size. For the element-wise operations, including Add, Mul, Neg, Abs, Inv, Sqrt, Square, Exp, and Log, the output dimension is the same as the input dimension. The aggregation operations, including Max, Sum, and Softmax, are performed on the second dimension and keep that dimension without reduction. It is worth noting that for each expression coded by a tree, the output $o$ is a matrix of shape $(N, C)$ or $(N, 1)$, so it needs to be calculated to a real number by:
\begin{equation}
\ell_s = \frac{1}{N} {\sum_{n=1}^{N} {\sum_{c=1}^{C\text{ or }1} {o_{nc}}}} \text{.}
\label{equ:fix_ope}
\end{equation}
A surrogate loss comprises a tree-encoded expression and fixed aggregation operations in Equation \ref{equ:fix_ope}.

\begin{table}[t] 
\centering 
\caption{Primitive operation set. The shape of $a$ and $b$ is $(N, C)$ or $(N, 1)$, where $N$ denotes the batch size and $C$ denotes the number of categories. $\gamma$ is a small constant to avoid dividing by zero.} \label{tab:ope_set}
\begin{tabular}{l|l|c} 
\toprule
Operation & Expression                            & Arity \\ \midrule
Add       & $a + b$                               & 2     \\
Mul       & $a \times b$                          & 2     \\ 
Neg       & $-a$                                  & 1     \\
Abs       & $|a|$                                 & 1     \\
Inv       & $\sign(a) / (|a| + \gamma)$           & 1     \\
Sqrt      & $\sign(a) \times \sqrt{|a| + \gamma}$ & 1     \\
Square    & $a^2$                                 & 1     \\
Exp       & $\exp(a)$                             & 1     \\
Log       & $\sign(a) \times \log(|a| + \gamma)$  & 1     \\ \midrule
Max       & $\max(a)$                             & 1     \\
Sum       & $\ssum(a)$                            & 1     \\
Softmax   & $\softmax(a)$                         & 1     \\
\bottomrule
\end{tabular}
\end{table}

\subsubsection{Search Algorithm}
We adopt GP, a population-based algorithm, to find a suitable surrogate loss to tighten the approximation error in the search space. As shown in Figure \ref{fig:pip_alar}, the search algorithm consists of 5 steps: initialization, evaluation, selection, crossover, and mutation.

\textit{1) Initialization:} In the initialization step, a number of expression trees are randomly generated to form the initial population.

\begin{figure}[t]
\centering
\includegraphics[width=8.0cm]{\impath/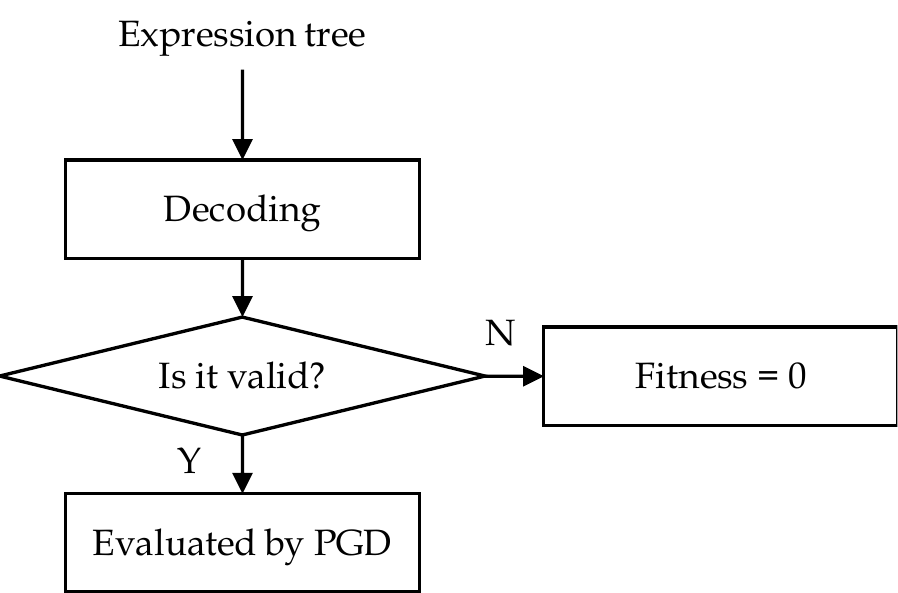}
\caption{Fitness evaluation procedure used in AutoLoss-AR.} 
\label{fig:fit_eva}
\end{figure}

\textit{2) Evaluation:} The fitness evaluation procedure is shown in Figure \ref{fig:fit_eva}. During the evaluation process, the generated expressions may be infeasible to compute, such as invalid values NaN, Inf, etc. At this point, we set their fitness to 0. Since the optimization objective is Equation \ref{equ:fit}, we directly calculate $R^{\prime\prime}$ for each valid tree as its fitness, where the value is between 0 and 1. The calculation of $R^{\prime\prime}$ is shown in Equation \ref{equ:rpp}, and we choose PGD-10 \cite{madry2017towards} as the algorithm $m$. To further reduce the time consumption, the approximate risk of an expression is calculated over 1000 images during the search. When testing the losses discovered, we calculate the approximate risk on the entire test set.

\textit{3) Selection:} We adopt the tournament \cite{goldberg1991comparative} as the selection strategy in AutoLoss-AR. It involves holding several tournaments among a few individuals randomly chosen from the population. The winner of each tournament, i.e., the one with the maximum fitness, is selected for the next step. Two hyperparameters, i.e., the size of the tournament and the number of times it is held, are used to control the selection process.

\textit{4) Crossover and Mutation:} The crossover and mutation steps are used to generate the offspring population. These expression trees are randomly subjected to one-point crossover and independent mutation, where the probabilities of mating and mutating are two hyperparameters.

\section{Experiments} \label{sec:exp}
\subsection{Purpose}
The purpose of our experiments is to answer:
\begin{itemize}
\item Under given conditions, i.e., $f$, $D$, $B$, and $\epsilon$, can AutoLoss-AR obtain a better solution than the handcrafted losses for tightening the approximation error?
\item Can the best-searched losses be used in other conditions, such as an \textit{unseen} model $f$?
\item if the searched surrogate losses perform better than the handcrafted baselines, then why?
\end{itemize}

\begin{table*}[t] 
\centering 
\caption{Adversarial accuracy of 10 adversarially trained models in white-box setting. The $\text{PGD}_{\text{BS}}$ column is the results of the best-searched losses with PGD-100. The Diff. column is the differences between the best-handcrafted losses and the best-searched losses, where the negative values indicate that the searched losses perform better than the handcrafted baselines.} \label{fig:res_whi_box}
\begin{tabular}{l||c||ccccc||c||c} 
\toprule
Methods & Clean & \PGDCE & \PGDCW & \APGDCE & \APGDCW & \APGDDLR & \PGDBS & Diff. \\ \midrule
\multicolumn{8}{l}{} \\
\multicolumn{8}{l}{CIFAR-10, $L_{\infty}$, $\epsilon = 8 / 255$} \\ \midrule
Addepalli2021 \cite{addepalli2021towards} & 80.24 & 55.78 & 51.70 & 55.78 & 51.72 & 51.75 & \underline{51.31} & \textbf{-0.39} \\
Cui2021 \cite{cui2021learnable}           & 88.22 & 53.97 & 53.77 & 53.89 & 53.75 & 56.05 & \underline{52.79} & \textbf{-0.96} \\
Huang2020 \cite{huang2020self}            & 83.48 & 55.92 & 53.95 & 55.83 & 53.97 & 54.42 & \underline{53.10} & \textbf{-0.85} \\
Sehwag2021\cite{sehwag2021improving}      & 84.38 & 57.42 & 56.43 & 57.30 & 56.37 & 56.92 & \underline{54.76} & \textbf{-1.61} \\
Wu2020\cite{wu2020adversarial}            & 85.36 & 58.80 & 56.82 & 58.86 & 56.80 & 56.90 & \underline{56.32} & \textbf{-0.48} \\
Zhang2019 \cite{zhang2019theoretically}   & 84.92 & 54.91 & 53.58 & 54.79 & 53.54 & 53.69 & \underline{52.94} & \textbf{-0.60} \\
\multicolumn{8}{l}{} \\
\multicolumn{8}{l}{CIFAR-10, $L_2$, $\epsilon = 0.5$} \\  \midrule
Sehwag2021 \cite{sehwag2021improving}     & 89.52 & 74.13 & 74.09 & 74.15 & 74.15 & 74.26 & \underline{73.48} & \textbf{-0.60} \\
Wu2020\cite{wu2020adversarial}            & 88.51 & 74.74 & 73.91 & 74.73 & 74.73 & 73.91 & \underline{73.69} & \textbf{-0.22} \\
\multicolumn{8}{l}{} \\
\multicolumn{8}{l}{CIFAR-100, $L_{\infty}$, $\epsilon = 8 / 255$} \\  \midrule
Addepalli2021 \cite{addepalli2021towards} & 62.02 & 32.83 & 27.98 & 32.87 & 27.95 & 28.05 & \underline{27.36} & \textbf{-0.59} \\
Cui2021 \cite{cui2021learnable}           & 70.25 & 30.13 & 28.20 & 29.90 & 28.10 & 29.59 & \underline{27.16} & \textbf{-0.94} \\
\bottomrule
\end{tabular}
\end{table*}

\subsection{Setup}
To test the performance of AutoLoss-AR and the searched surrogate losses, we evaluate the robustness of 17 adversarially trained models\footnote{All the adversarially trained models used in this paper are obtained from \url{https://robustbench.github.io}.}, using CIFAR-10 \cite{krizhevsky2009learning} and CIFAR-100 \cite{krizhevsky2009learning} as datasets. Of these, 10 models are used for white-boxing testing, i.e., searching for losses directly on these models using AutoLoss-AR. The other 7 models are used for black-box testing, i.e., their robustness is evaluated using the previously searched functions.

Both searching for surrogate losses and using the searched losses for adversarial robustness evaluation involve the selection of an algorithm $m$. In the search process of AutoLoss-AR, considering the time consumption, we choose 10-step PGD (PGD-10) \cite{madry2017towards} as the algorithm $m$. Once the search is completed, the best-searched losses are combined with 100-step PGD (PGD-100) \cite{madry2017towards} and 100-step Auto-PGD (APGD-100) \cite{croce2020reliable} to evaluate the adversarial robustness of a model.

\begin{table}[t] 
\centering 
\caption{Hyperparameters of AutoLoss-AR used in this paper.} \label{tab:hyp}
\begin{tabular}{l|c} 
\toprule
Hyperparameter          & Value \\ \midrule
Number of generations   & 50    \\
Maximum depth of a tree & 25    \\
Population size         & 100   \\
Tournament size         & 3     \\
Crossover rate          & 0.5   \\
Mutation rate           & 0.3   \\
\bottomrule
\end{tabular}
\end{table}

To provide a comparison, CE loss \cite{goodfellow2014explaining,madry2017towards,dong2018boosting}, CW loss \cite{carlini2017towards}, and DLR loss \cite{croce2020reliable} are selected as the handcrafted baselines. Carlini and Wagner \cite{carlini2017towards} tested 7 different losses, and here we choose only the one with the best performance. Since our goal is to find a surrogate loss suitable for adversarial robustness evaluation, the most straightforward measure for the performance of a loss is the approximate risk evaluated using this function, i.e., $R^{\prime\prime}$. However, to be consistent with previous studies \cite{goodfellow2014explaining,madry2017towards,carlini2019evaluating,croce2020reliable}, we use adversary accuracy as the measure, which is equal to $1 - R^{\prime\prime}$. 

AutoLoss-AR is based on tree coding and GP algorithm. Some hyperparameters of AutoLoss-AR used in this paper are shown in Table \ref{tab:hyp}. Our experiments are implemented with PyTorch \cite{paszke2017automatic} and DEAP \cite{de2012deap}, and run on 4 Tesla V100 GPUs.

\begin{table*}[t] 
\centering 
\caption{Formulas of the simplified best-searched surrogate losses.} \label{tab:for_los}
\begin{tabular}{l|l} 
\toprule
Searched Loss & Formula                                                                                     \\ \midrule
BS1           & $\exp(10 \times \softmax(p) / \max(\softmax(p)))$                                           \\
BS2           & $\exp(- \max(\softmax(p + 2 \times \softmax(5 \times p))))$                                 \\ 
BS3           & $\softmax(-\softmax(\exp(p) \times 2 \times p)) \times (\softmax(2 \times p) + 2 \times q)$ \\
BS4           & $(\softmax(\softmax(2 \times p) + p - q) - q)^2$                                            \\
BS5           & $\exp(- \max(\softmax(\exp(\softmax(\exp(p) + p) + 1) + p) + 1))$                           \\
\bottomrule
\end{tabular}
\end{table*}

\subsection{White-box Results}
Under given $f$, $D$, $B$, and $\epsilon$, can AutoLoss-AR obtain a better solution than the handcrafted losses for tightening the approximation error? To answer this question, we use the proposed method to find surrogate losses on 10 adversarially trained models. The results are shown in Table \ref{fig:res_whi_box}. As we can see, the adversarial accuracy evaluated using the searched losses outperforms the accuracy evaluated using the baselines for all models. The improvements are around 0.2\% to 1.6\%. These results suggest that the answer to the question is: yes, AutoLoss-AR can obtain better surrogate losses than the manual design baselines for closer approximations of the true adversarial risk. 

\begin{figure}[t]
\centering
\includegraphics[width=8.0cm]{\impath/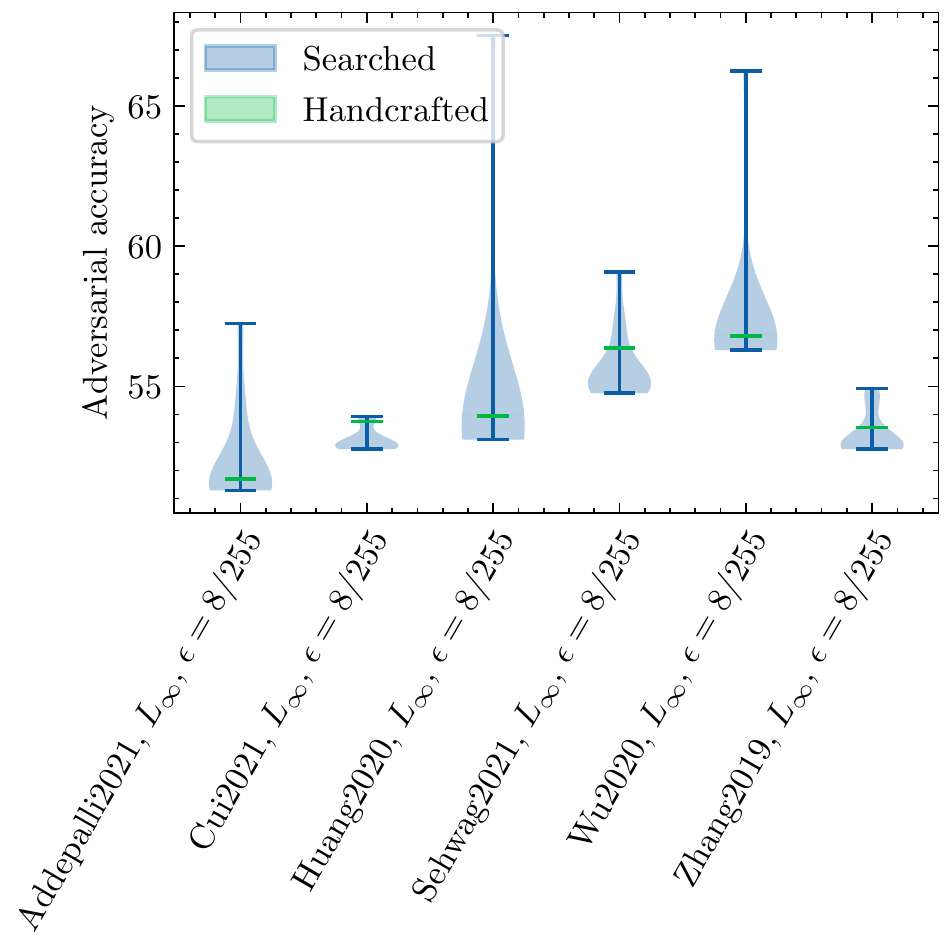}
\caption{Adversarial accuracy of the losses searched by AutoLoss-AR in 15 independent runs. The green lines represent the best-handcrafted baselines.} 
\label{fig:sea_suc_rat}
\end{figure}

Another point of interest is the search efficiency of AutoLoss-AR. We perform 15 independent searches with the proposed method on six models trained using CIFAR-10 and $L_{\infty}$-norm adversarial training, and the results are shown in Figure \ref{fig:sea_suc_rat}. The run rates that can search for surrogate losses better than the best-handcrafted baselines are 9/15, 12/15, 12/15, 12/15, 8/15, 12/15 for the six models, respectively.

\subsection{Black-box Results}
Can the searched loss function be used in other conditions? This problem is of practical interest because it determines when evaluating the adversarial robustness of a new model, whether the search losses can be used independently as some new knowledge without going through a time-consuming search process. We try to figure it out in this section.

First, we simplify the searched functions in white-box setting and extract 5 clear and readable formulas, as shown in Table \ref{tab:for_los}. It can be seen that these functions have some common features, i.e., they frequently use $\exp$,  $\max$, and $\softmax$ as the operator units. Then, these losses are used to evaluate the adversarial robustness of 7 \textit{unseen} adversarially trained models, and the results are shown in Figure \ref{fig:res_bla_box}. 

Some observations are summarized as follows:
\begin{itemize}
\item The searched losses performed better than the handcrafted baselines in the vast majority of cases. Of the 210 adversarial accuracies evaluated using the searched losses, only 3 performed worse than the manual ones. The best-discovered losses outperform the best baselines by 0.8\% to 2.4\%.
\item BS5, BS3, and BS2 are the top 3 best-performing losses among these 5 simplified functions. BS5 and BS2 have similar formulas, as shown in Table \ref{tab:for_los}, where they both contain $\exp(-\max(\softmax(\cdots)))$.
\item The loss functions searched by using PGD-10 as $m$ can also be well suitable for PGD-100 and APGD-100.
\item The loss functions searched by using $8/255$ as $\epsilon$ can also be well suitable for $10/255$ and $12/255$.
\end{itemize} 

The above results and observations illustrate that the simplified 5 losses can be used individually as some kind of new knowledge. The robustness of the model evaluated using them is in most cases more accurate than the hand-designed baselines.

\begin{sidewaystable*}
\centering \tiny
\caption{Adversarial accuracy of 7 adversarially trained models in black-box setting. The $\text{PGD}_{\text{BSi}}$ column is the results of the BSi loss with PGD-100, where i = 1, 2, 3, 4, 5. The $\text{APGD}_{\text{BSi}}$ column is the results of the BSi loss with APGD-100. The Diff. column is the differences between the best-handcrafted losses and the best-searched losses, where the negative values indicate that the searched losses perform better than the handcrafted baselines.} \label{fig:res_bla_box}
\begin{tabular}{l||c||ccccc||ccccc||ccccc||c} 
\toprule
Methods & Clean & \PGDCE & \PGDCW & \APGDCE & \APGDCW & \APGDDLR & \PGDBSon & \PGDBStw & \PGDBSth & \PGDBSfo & \PGDBSfi& \APGDBSon & \APGDBStw & \APGDBSth & \APGDBSfo & \APGDBSfi & Diff. \\ \midrule
\multicolumn{8}{l}{} \\
\multicolumn{8}{l}{CIFAR-10, $L_{\infty}$, $\epsilon = 8 / 255$} \\ \midrule
Carmon2019 \cite{carmon2019unlabeled}   & 89.69 & 61.87 & 60.60 & 61.85 & 60.64 & 60.87 & 59.95 & 59.80 & 59.90 & 59.82 & 59.84 & 59.98 & 59.82 & 59.82 & 59.81 & \underline{59.80} & \textbf{-0.80} \\
Engstrom2019 \cite{engstrom4robustness} & 87.03 & 51.85 & 52.40 & 51.77 & 52.34 & 53.11 & 50.64 & 50.11 & 50.42 & 50.18 & 50.17 & 50.65 & \underline{50.07} & 50.41 & 50.17 & 50.12 & \textbf{-1.70} \\
Hendrycks2019 \cite{hendrycks2019using} & 87.11 & 57.25 & 56.54 & 57.20 & 56.39 & 57.18 & 55.30 & 55.20 & 55.21 & 55.22 & 55.17 & 55.27 & 55.20 & \underline{55.13} & 55.16 & 55.17 & \textbf{-1.26} \\
Rice2020 \cite{rice2020overfitting}     & 85.34 & 57.01 & 55.38 & 56.92 & 55.33 & 56.04 & 54.13 & 53.87 & 53.81 & 53.90 & 53.83 & 54.15 & 53.81 & \underline{53.80} & 53.88 & 53.86 & \textbf{-1.53} \\
Wang2019 \cite{wang2019improving}       & 87.50 & 61.81 & 58.30 & 61.74 & 58.12 & 58.95 & 57.43 & 57.54 & 57.30 & 57.80 & 57.48 & 57.35 & 57.54 & \underline{57.23} & 57.73 & 57.31 & \textbf{-0.89} \\
Wong2020 \cite{wong2020fast}            & 83.34 & 46.01 & 45.96 & 45.91 & 45.95 & 47.15 & 44.17 & 43.65 & 43.73 & 43.76 & 43.67 & 44.09 & \underline{43.61} & 43.72 & 43.72 & 43.62 & \textbf{-2.30} \\
Zhang2020 \cite{zhang2020attacks}       & 84.52 & 56.78 & 54.50 & 56.78 & 54.51 & 54.77 & 53.77 & 53.82 & 53.74 & 53.92 & 53.77 & 53.73 & 53.81 & \underline{53.72} & 53.88 & 53.73 & \textbf{-0.78} \\
\multicolumn{8}{l}{} \\
\multicolumn{8}{l}{CIFAR-10, $L_{\infty}$, $\epsilon = 10 / 255$} \\  \midrule
Carmon2019 \cite{carmon2019unlabeled}   & 89.69 & 51.19 & 50.41 & 51.00 & 50.50 & 50.88 & 49.25 & 49.03 & 49.12 & 49.01 & 49.00 & 49.25 & \underline{48.89} & 48.97 & 48.93 & 48.92 & \textbf{-1.52} \\
Engstrom2019 \cite{engstrom4robustness} & 87.03 & 40.08 & 41.28 & 39.80 & 41.09 & 42.53 & 39.27 & 38.60 & 39.05 & 38.80 & 38.62 & 39.14 & 38.38 & 38.86 & 38.66 & \underline{38.36} & \textbf{-1.44} \\
Hendrycks2019 \cite{hendrycks2019using} & 87.11 & 47.41 & 46.69 & 47.33 & 46.55 & 47.70 & 44.94 & 44.64 & 44.76 & 44.60 & 44.79 & 44.82 & 44.59 & 44.67 & \underline{44.57} & 44.67 & \textbf{-1.98} \\
Rice2020 \cite{rice2020overfitting}     & 85.34 & 46.82 & 45.53 & 46.72 & 45.38 & 46.48 & 43.83 & 43.53 & 43.65 & 43.54 & 43.64 & 43.84 & 43.49 & 43.56 & \underline{43.42} & 43.52 & \textbf{-1.96} \\
Wang2019 \cite{wang2019improving}       & 87.50 & 52.35 & 49.05 & 52.00 & 48.89 & 50.12 & 47.81 & 47.81 & 47.59 & 48.05 & 47.83 & 47.62 & 47.74 & 47.59 & 47.93 & \underline{47.55} & \textbf{-1.34} \\
Wong2020 \cite{wong2020fast}            & 83.34 & 35.28 & 35.94 & 35.22 & 35.86 & 37.40 & 33.81 & 33.38 & 33.60 & 33.46 & 33.32 & 33.75 & 33.25 & 33.53 & 33.34 & \underline{33.14} & \textbf{-2.08} \\
Zhang2020 \cite{zhang2020attacks}       & 84.52 & 48.32 & 46.30 & 48.25 & 46.20 & 46.57 & 45.18 & 45.00 & 44.93 & 45.12 & 44.99 & 45.13 & 44.97 & \underline{44.92} & 45.02 & 44.95 & \textbf{-1.28} \\
\multicolumn{8}{l}{} \\
\multicolumn{8}{l}{CIFAR-10, $L_{\infty}$, $\epsilon = 12 / 255$} \\  \midrule
Carmon2019 \cite{carmon2019unlabeled}   & 89.69 & 40.97 & 41.01 & 40.73 & 40.81 & 41.33 & 39.60 & 38.85 & 39.28 & 38.94 & 39.02 & 39.41 & 38.86 & 39.16 & 38.91 & \underline{38.84} & \textbf{-1.89} \\
Engstrom2019 \cite{engstrom4robustness} & 87.03 & 29.65 & 31.69 & 29.43 & 31.38 & 32.96 & 29.55 & 28.83 & 29.54 & 29.12 & 28.79 & 29.50 & 28.69 & 29.32 & 29.02 & \underline{28.56} & \textbf{-0.87} \\
Hendrycks2019 \cite{hendrycks2019using} & 87.11 & 36.52 & 37.05 & 36.37 & 36.86 & 38.34 & 34.97 & 34.30 & 34.65 & 34.55 & 34.36 & 34.82 & \underline{34.24} & 34.58 & 34.40 & 34.25 & \textbf{-2.13} \\
Rice2020 \cite{rice2020overfitting}     & 85.34 & 36.14 & 35.99 & 36.14 & 35.80 & 37.15 & 34.08 & 33.52 & 33.90 & 33.65 & 33.49 & 33.92 & 33.44 & 33.70 & 33.46 & \underline{33.43} & \textbf{-2.37} \\
Wang2019 \cite{wang2019improving}       & 87.50 & 42.20 & 39.21 & 41.89 & 39.12 & 40.53 & 38.01 & 38.18 & 38.03 & 38.32 & 38.20 & 37.89 & 37.97 & \underline{37.72} & 38.07 & 37.91 & \textbf{-1.40} \\
Wong2020 \cite{wong2020fast}            & 83.34 & 25.24 & 26.75 & 25.04 & 26.54 & 28.14 & 24.62 & 23.71 & 24.27 & 23.98 & 23.86 & 24.42 & 23.63 & 24.10 & 23.93 & \underline{23.62} & \textbf{-1.42} \\
Zhang2020 \cite{zhang2020attacks}       & 84.52 & 37.81 & 54.50 & 39.51 & 37.64 & 37.95 & 36.67 & 36.53 & 36.51 & 36.55 & 36.43 & 36.53 & 36.34 & 36.37 & \underline{36.33} & 36.37 & \textbf{-1.31} \\
\bottomrule
\end{tabular}
\end{sidewaystable*}

\subsection{Attribution Study}
\begin{figure*}[t]
\centering
\subfigure[CE loss, BS3 loss]{\includegraphics[width=8.5cm]{\impath/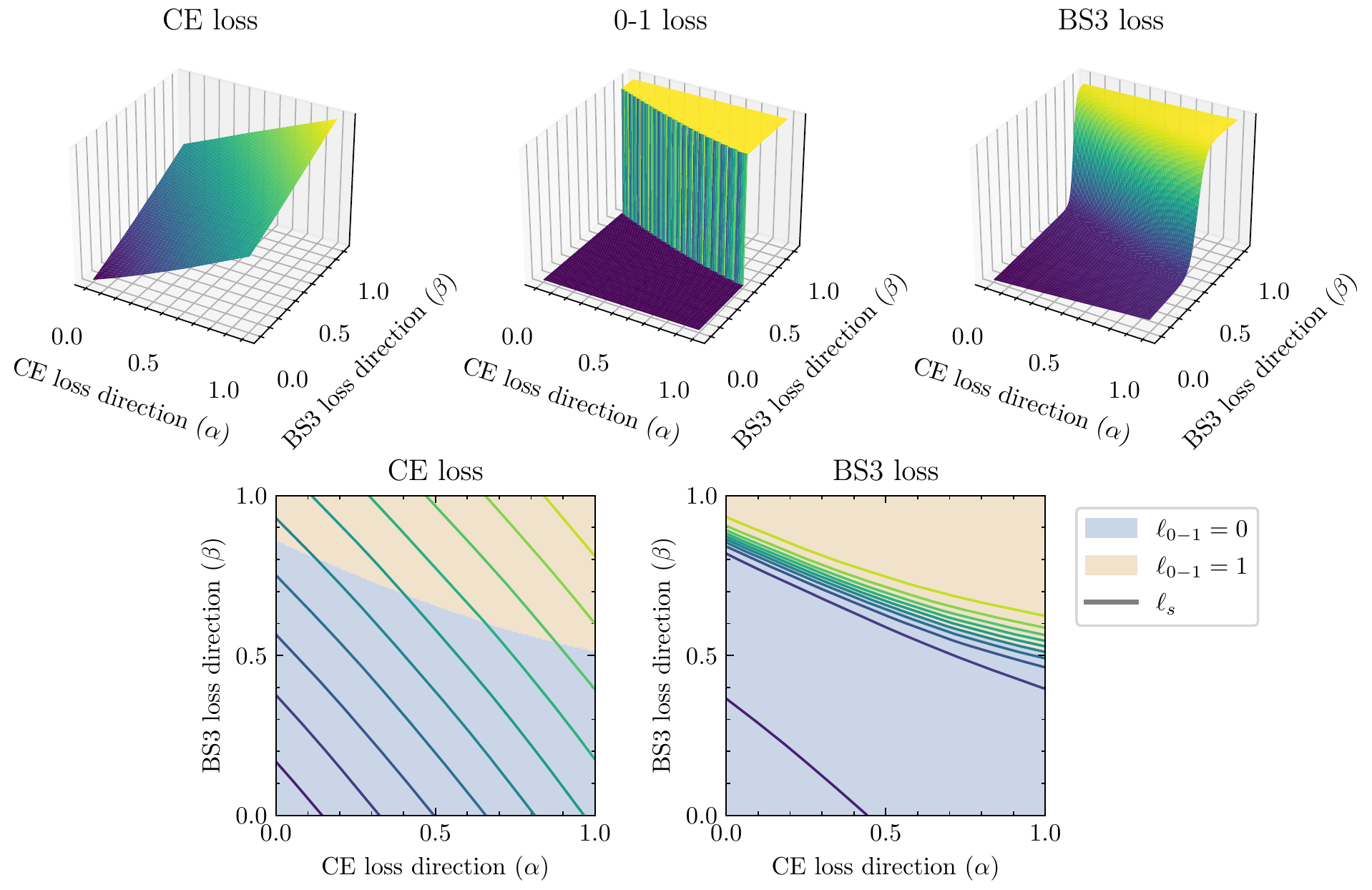} \label{fig:los_lan_a}}
\subfigure[CW loss, BS3 loss]{\includegraphics[width=8.5cm]{\impath/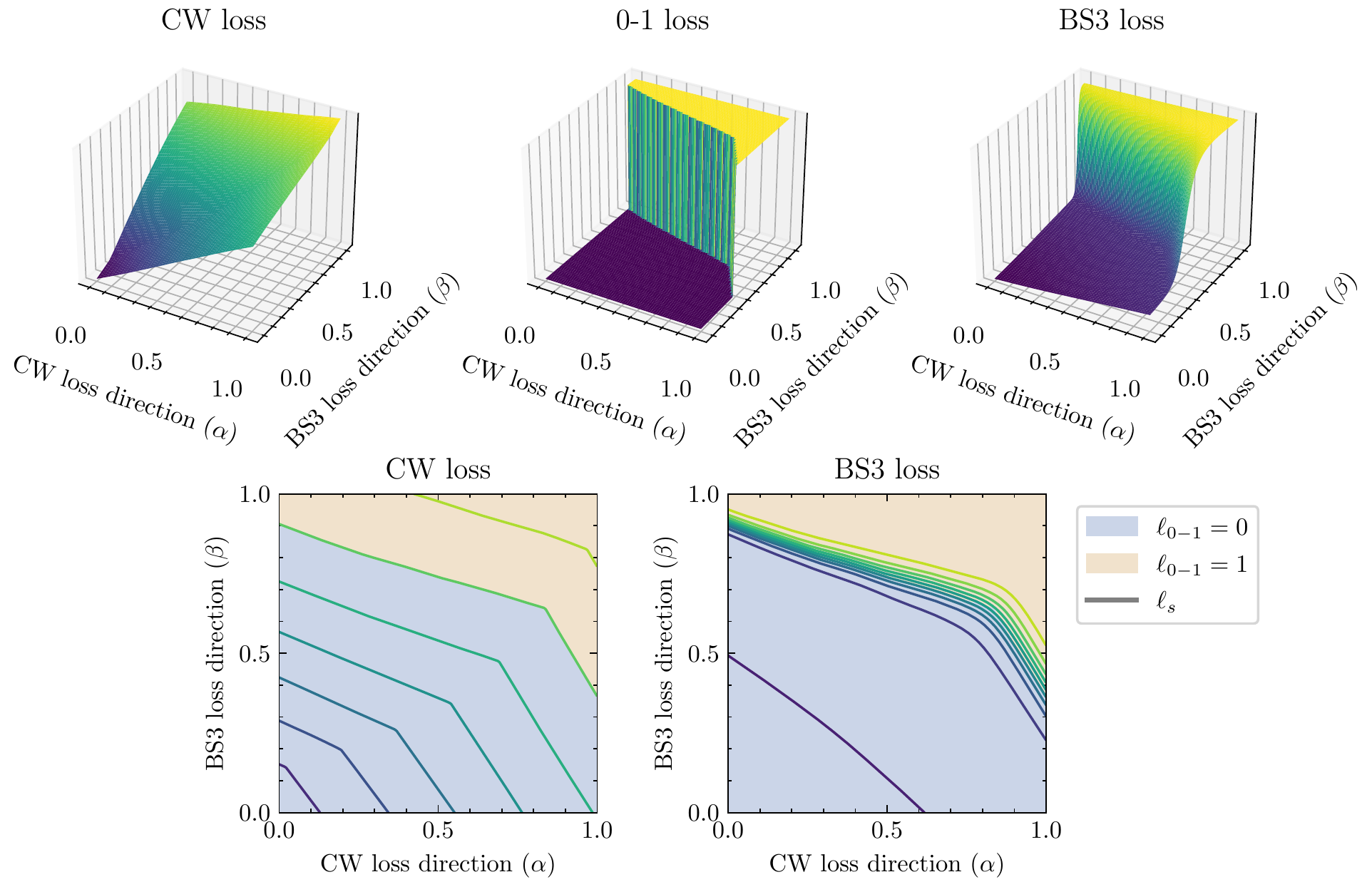} \label{fig:los_lan_b}}
\subfigure[CE loss, BS5 loss]{\includegraphics[width=8.5cm]{\impath/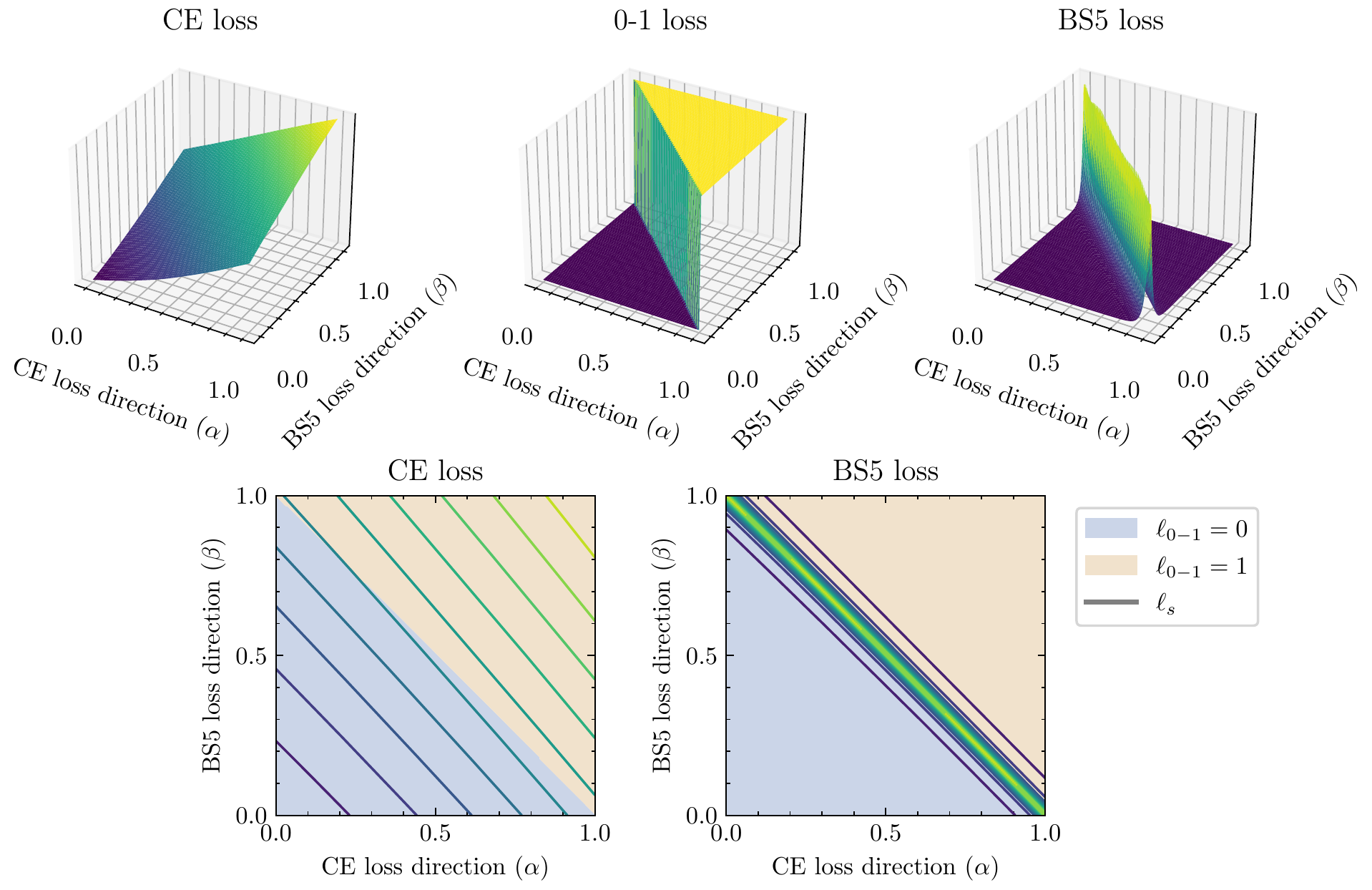} \label{fig:los_lan_c}}
\subfigure[CW loss, BS5 loss]{\includegraphics[width=8.5cm]{\impath/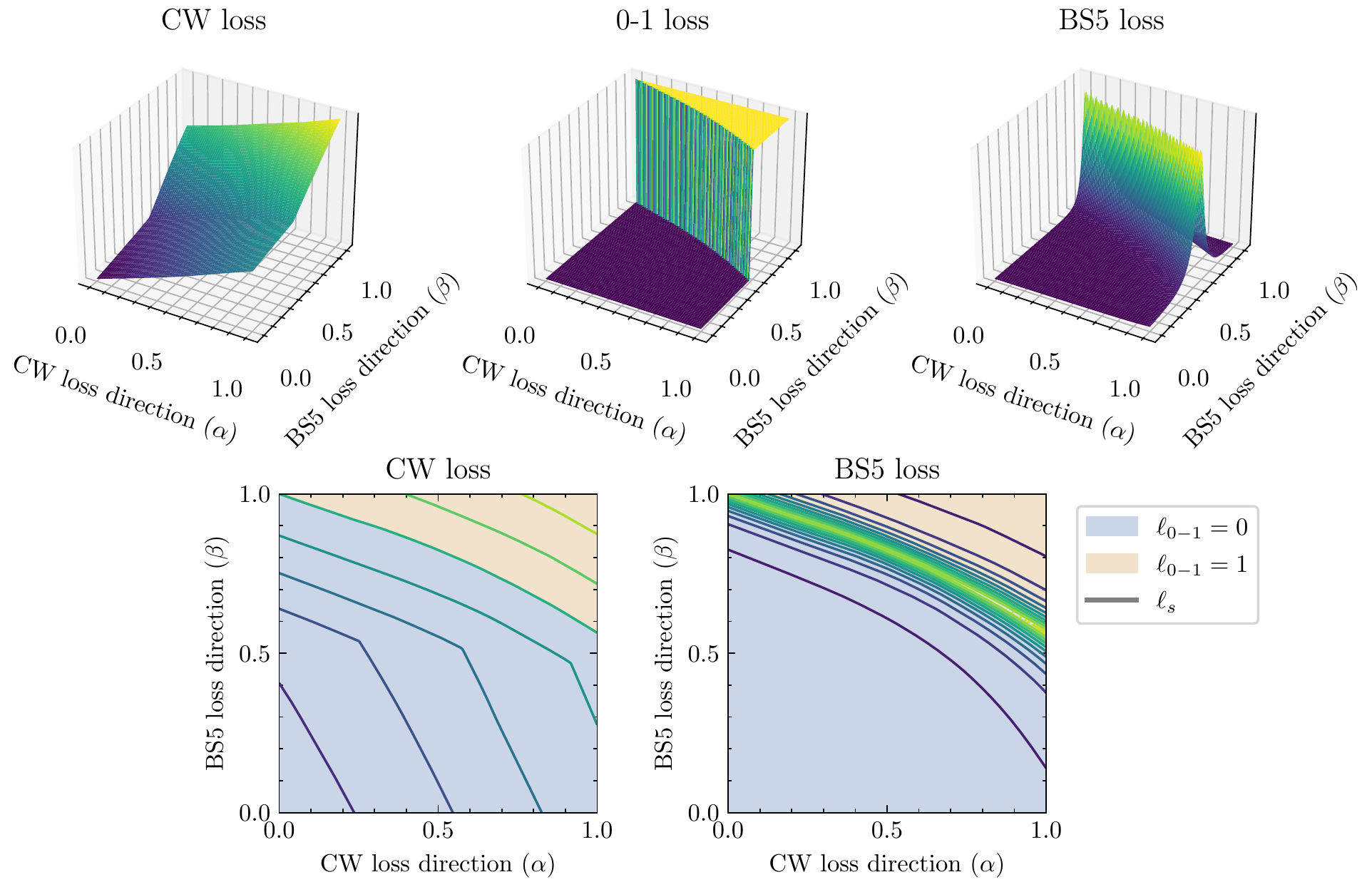} \label{fig:los_lan_d}}
\caption{Local loss landscapes on Wong2020 \cite{wong2020fast}.} 
\label{fig:los_lan}
\end{figure*}

If the searched surrogate losses perform better than the handcraft ones, then why? To answer this question, we visualize the local loss landscapes, as shown in Figure \ref{fig:los_lan}. We first generate the adversarial examples, i.e., $x^{\prime}_{\text{HC}}$ and $x^{\prime}_{\text{BS}}$ of the clean input $x$ by the handcraft losses and the searched losses, respectively, and keep $f(x^{\prime}_{\text{HC}}) = y$ and $f(x^{\prime}_{\text{BS}}) \neq y$. Then, we calculate the loss values by:
\begin{equation}
\ell(\alpha, \beta) = \clamp_{0, 1}(\alpha \times (x^{\prime}_{\text{HC}} - x) + \beta \times (x^{\prime}_{\text{BS}} - x)) \text{,}
\end{equation}
where $0 \le \alpha \le 1$ and $0 \le \beta \le 1$. In Figure \ref{fig:los_lan}, $\ell_{0-1}(\alpha, \beta)$, $\ell_{\text{CE}}(\alpha, \beta)$, $\ell_{\text{CW}}(\alpha, \beta)$, $\ell_{\text{BS3}}(\alpha, \beta)$, and $\ell_{\text{BS5}}(\alpha, \beta)$ are drawn.

The results indicate that there may be two reasons. The first one is the local consistency between the searched loss and the 0-1 loss is better than the handcrafted loss. For example, as shown in Figure \ref{fig:los_lan_a}, maximizing CE loss does not maximize the 0-1 loss. The second one is the specific algorithm $m$, such as PGD-100 used here, cannot guarantee to find the global optimum, although the handcrafted loss has good consistency with the 0-1 loss, such as Figure \ref{fig:los_lan_b}. The searched surrogate is easier for the algorithm to find a more suitable adversarial example. Interestingly, we find that both BS3 and BS5 are doing their best to smoothly simulate the steep 0-1 loss (for BS5, the part that rises from $\alpha=0$ and $\beta=0$ looks like this). Perhaps such losses are better fit with the algorithm $m$ and easier to find adversarial examples.

\subsection{Qualitative Results}
\begin{figure*}[h]
\centering
\includegraphics[width=17.0cm]{\impath/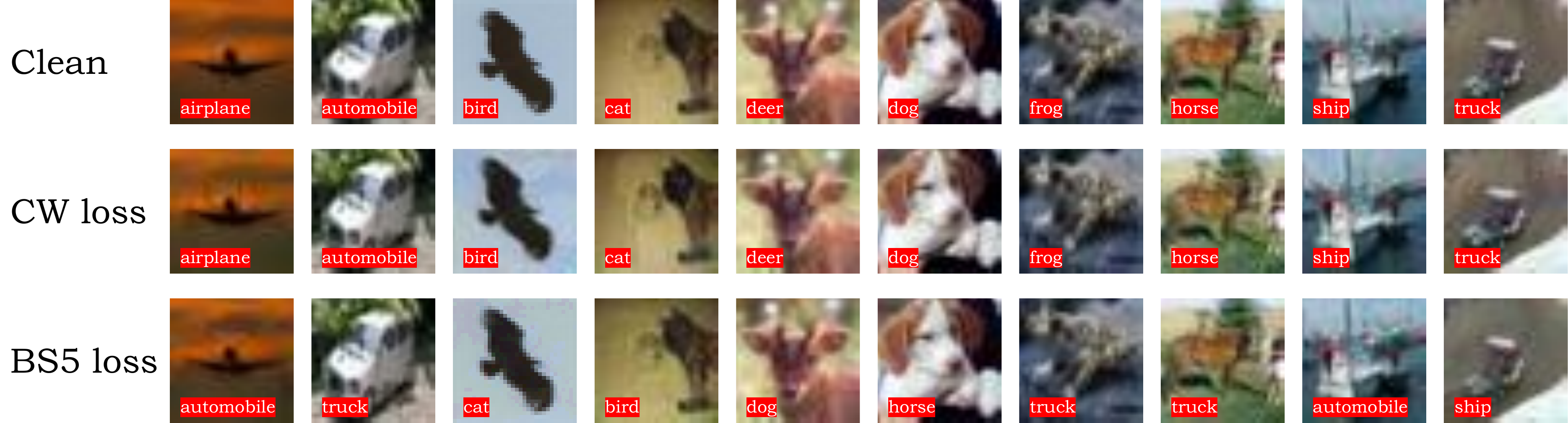}
\caption{Qualitative examples of adversarial examples generated with CW loss and BS5 loss.} 
\label{fig:adv_exa}
\end{figure*}

Some qualitative examples are shown in Figure \ref{fig:adv_exa}, where the adversarial examples generated using BS5 loss can misled the model and are visually similar to clean images.

\section{Conclusion} \label{sec:con}
We establish the tightening of the approximation error as an optimization problem and solve it with AutoML. Specifically, we focus on the surrogate loss in adversarial robustness evaluation and propose AutoLoss-AR to find a possible solution with tree coding and GP algorithm. The experiments on 10 adversarially trained models are conducted to verify the effectiveness and efficiency of the proposed method. The results show that the risks evaluated using the best-discovered losses are 0.2\% to 1.6\% better than those evaluated using the handcrafted baselines. Meanwhile, 5 surrogate losses with clean and readable formulas are distilled out. The experimental results demonstrate that they perform well on \textit{unseen} adversarially trained models. Besides, we also identify two reasons why the searched losses are better than the baselines by visualizing the local loss landscapes: (1) the local consistency between the searched loss and the 0-1 loss is better, and (2) the searched loss is easier to optimize to find a deceptive adversarial example.

\section*{Acknowledgment}
The work was supported in part by the National Natural Science Foundation of China under Grands U19B2044 and 61836011.

\bibliographystyle{ACM-Reference-Format}
\bibliography{./references.bib}

\end{document}